\documentclass{article}

\usepackage{graphics}
\usepackage{epsfig}
\usepackage{float}

\usepackage[accepted]{icml2008}

\usepackage{graphicx} 

\usepackage{mlapa}

\icmltitlerunning{Conditional anomaly detection methods for patient--management 
alert systems}

\begin{document} 

\twocolumn[
\icmltitle{Conditional anomaly detection methods \\ 
for patient-management 
alert systems}

 \icmlauthor{Michal Valko}{michal@cs.pitt.edu}
 \icmlauthor{Gregory Cooper}{gfc@cbmi.pitt.edu}
 \icmlauthor{Amy Seybert}{seyberta@upmc.edu}
 \icmlauthor{Shyam Visweswaran}{shv3@pitt.edu}
 \icmlauthor{Melissa Saul}{saulmi@upmc.edu}
 \icmlauthor{Milo\v{s} Hauskrecht}{milos@cs.pitt.edu}
 \icmladdress{University of Pittsburgh, PA}

\vskip 0.3in
]

\newcommand{\A}{{\cal A}}
\newcommand{\coA}{\bar{A}}
\newcommand{\B}{{\cal B}}
\newcommand{\E}{{\cal E}}
\newcommand{\bX}{{\bf X}}
\newcommand{\bY}{{\bf Y}}
\newcommand{\bZ}{{\bf Z}}
\newcommand{\bW}{{\bf W}}
\newcommand{\bC}{{\bf C}}
\newcommand{\bB}{{\bf B}}
\newcommand{\bA}{{\bf A}}
\newcommand{\bx}{{\bf x}}
\newcommand{\ba}{{\bf a}}
\newcommand{\bb}{{\bf b}}
\newcommand{\bc}{{\bf c}}
\newcommand{\bv}{{\bf v}}
\newcommand{\bw}{{\bf w}}
\newcommand{\bu}{{\bf u}}
\newcommand{\by}{{\bf y}}
\newcommand{\bz}{{\bf z}}
\newcommand{\bq}{{\bf q}}
\newcommand{\bt}{{\bf t}}
\newcommand{\vecb}{{\bf b}}
\newcommand{\bT}{{\bf T}}

\newcommand{\mynote}[1]{\begin{center}\fbox{\parbox{5in}{#1}}\end{center}}
\newcommand{\commentout}[1]{}
\newcommand{\highl}[1]{\em {#1}}
\newcommand{\action}[0]{\eta}
\newcommand{\vect}[1]{\bf {#1}}
\newcommand{\real}{\hbox{\it I\hskip -2pt R}}

\newcommand{\Naive}{Na\"{\i}ve{ }}
\newcommand{\naive}{na\"{\i}ve{ }}

\begin{abstract} 
We develop and evaluate a data--driven approach for detecting unusual (anomalous) patient--management decisions using past patient cases stored in an electronic health record (EHR) system.. Our hypothesis is that patient--managements decisions that are unusual with respect to past patients may be due to potential error and that it is worthwhile to raise an alert if such a condition is encountered. We propose and develop two methods for detecting anomalies that are based on predictive classification models used in supervised learning applications. We evaluate the new approach by identifying patients who are at risk Heparin induced Thrombocytopenia (HIT), a condition that is life--threatening if it is not detected and managed properly.  Our evaluations show that the detector based on our method is able to reach performance level superior to the existing screening method in terms of its precision and recall characteristics; hence demonstrating its potential utility in clinical applications. 
\end{abstract} 

\section{Introduction}
Anomaly detection methods can be very useful in identifying unusual or interesting patterns in data. A newly developed conditional anomaly detection framework \cite{hauskrecht2007evidence-based} extends anomaly detection to the problem of identifying anomalous patterns on a subset of attributes in the data. The anomaly always depends (is conditioned) on the value of remaining attributes. We believe such a framework is particularly useful in medicine in detection of unusual patient--management decisions that can be often associated with potential patient management errors. Consequently the new framework can be used to monitor and screen data for a broad range of potential patient errors and raise an alert if it finds one.
Our anomaly detection framework builds upon advances in classification model learning and prediction. Let $\bx$ defines a vector of context attributes (representing the patient's state) and $\by$ defines the target attribute (representing the target patient--management decision).  Our goal is to decide if the example ($\bx$,$\by$) is conditionally anomalous with respect to past examples (patients) in the database. In other words, we ask if the patient management decision $\by$ is unusual for the patient condition $\bx$, by taking into account records for past patients in the database. 
We propose an anomaly detection framework that solves the problem by building a measure $d( )$ that reflects the severity with which an example differs from conditional (context--to--target) patterns observed in the database. All anomaly calls are defined relative to this measure. To construct the model of d we build upon advances in classification model learning and prediction. In particular, our method exploits discriminant functions often used to make classification model calls and uses them to define the anomaly measure d.  We show how to design $d$ for the support vector machines \cite{vapnik1995nature}. The anomaly detection framework proposed in this work extends the probabilistic framework from \cite{hauskrecht2007evidence-based} and generalizes it to a much larger class of discriminative models.  
To demonstrate the benefits of our anomaly detection framework we test it on a problem of detecting test--order--decision anomalies in context of heparin--induced thrombocytopenia (HIT) \cite{warkentin2004heparin-induced} that occurs in approximately 2\% of population treated with heparin.  We show that our framework is able to detect anomalies created by inverting test--order--decision for the HPF4 antibody test with a very promising recall and precision characteristics. The HPF4 test is a test used to confirm the presence of the condition.

\section{Methodology}
The objective of ``standard" anomaly detection is to identify a data example a that deviates from all other examples $E$ in the database. Conditional anomaly detection \cite{hauskrecht2007evidence-based} is different. The goal is to detect an unusual pattern relating context attributes $\bx$ and target attributes $\by$ in the example $a$, that deviates from patterns observed in other examples in the database.   
To assess the conditional anomaly of a we propose to first build (learn) a one--dimensional projection $d(.)$  of data that reflects the prevailing (or expected) conditional pattern in the database for $\by$ given $\bx$. The projection model $d$ is then used to analyze the deviations of $a$'s to determine the anomaly. Briefly, we say that the case $a$ is anomalous in the target attribute(s) $\by$ with respect to context $\bx$, if the value $d(\by|\bx)$ differs from the values of other examples in the database. 
Our conditional anomaly detection framework can be used for a number of purposes. Our objective here is to use it detect anomalous patient--management decisions. In this case the context attributes x define the patient's condition and the target attribute y corresponds to the patient--management decision we want to evaluate. Our belief is that these conditional anomalies may often correspond to patient--management errors.   
In the following we present a method for building projections $d(.)$. This method is derived from the model used frequently in classification model learning: the support vector machines (SVM) \cite{vapnik1995nature}.  The fact that we use classification model is not a coincidence. Classification models attempt to learn conditional patterns in between inputs $\bx$ and class outputs $\by$ from past data and apply them to predict the class membership for future inputs.  In our case, we aim to model a relation in between context ($\bx$) and target patterns ($\by$) and apply it to detect pattern deviations in the new example $(\bx,\by)$. In both cases the model learning attempts to capture the prevailing conditional patterns observed in the dataset and the difference is in how the learned patterns are used in the two frameworks. 

\subsection{Anomaly detection}
One--dimensional discriminant projections defined above are the basis of anomaly calls made by our detector. Briefly, the anomaly is made if the value $d$ for a patient falls below a certain threshold. Since different models come with different projections the thresholds values need to be calibrated to the desired performance level, e.g. using acceptable specificity or false alarm rates.

\section{Experimental evaluation}

To demonstrate the benefits of our anomaly detection framework we test it on a problem of detecting test--order--decision anomalies in context of heparin--induced thrombocytopenia (HIT).  

\subsection{Heparin induced thrombocytopenia}

Heparin--induced thrombocytopenia (HIT) \cite{warkentin2004heparin-induced} is a transient pro--thrombotic disorder induced by heparin exposure with subsequent thrombocytopenia and associated thrombosis.   HIT is a condition that is life--threatening if it is not detected and managed properly.   Two tests are used to detect HIT: Heparin platelet antibody (HPA), and Heparin Platelet factor 4 antibody (HPF4) tests. The second test is more reliable and accurate. The importance of the HPF4 tests and its orders in the detection of the HIT condition makes the test an excellent candidate for the evaluation of our anomaly framework. 

\subsection{Data}
The HIT dataset used in our experiment was built from de--identified data selected from $3949$  records of post--surgical cardiac patients treated at one of the University of Pittsburgh Medical Center (UPMC) teaching hospitals. The data collected for patients were obtained from the MARS system, which serves as an archive for much of the data collected at UPMC. The records for individual patients included discharge records, demographics, all labs and tests (including standard and all special tests), two medication databases, and a financial charges database.  The data were cleaned and stored in a MySQL database.  For the purpose of this experiment the data were preprocessed and used to build a dataset of  $34589$ patient state examples for which the HPF4 test--order decision (order vs. no--order)  was considered and evaluated. The patient states were generated automatically at discrete time points marked by the arrival of a new platelet result, a key feature used in the HIT detection.  A total of $274$ HPF4 orders were associated with these states (prior of a test order is $0.79\%$)   Each data--point generated consisted of a total of $43$ features that included recent platelets, platelet trends, platelet drops from nadir and the first platelet value, a set of similar values for hemoglobin and hemoglobin trends, an indicator of the ongoing heparin treatment and the total time on heparin.   

\begin{figure}
 \begin{center}
      \includegraphics[width=8cm]{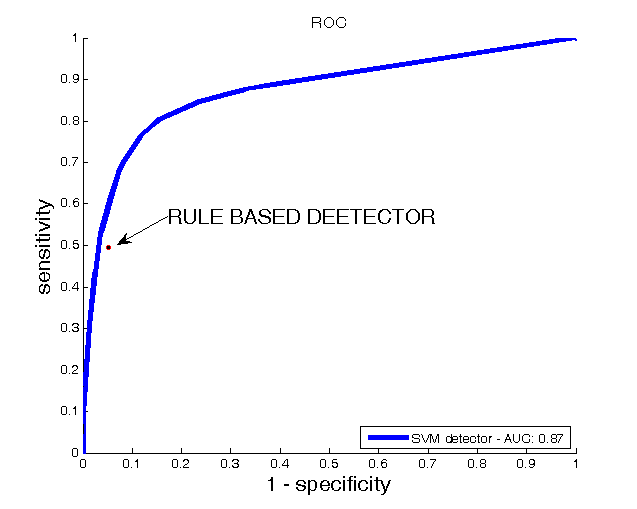}
      \caption{ROC curve for the anomaly detection method based on the and SVM projections on the HIT dataset. The methods are compared to the sensitivity and specificity of the rule based detector.}
      \label{fig:svm}
    \end{center}
    \end{figure}

\section{Results}
The comparisons of expected and detected anomalies under different detection thresholds were used to calculate sensitivities, specificities and the precision of the method.  The sensitivity and specificity pairs for the SVM projection methods are plotted on the ROC curve in Figure \ref{fig:svm}. The baseline for the comparison was the rule based detector described above. 
 
We see that SVM anomaly detector dominate the rule--based baseline with the specificity of 94\%, sensitivity of 49\%. Very importand summary statistics for the deployment however is positive predictive value (PPV) : How many alarms fired correspond to true alarms? For the baseline rule, PPV was 7.2\% (which corresponds to 136 true and 1752 false positives). With our SVM detector we were able roughly double PPV with reasonable number of true positives: 15.6\% (84 true and 453 false positives).

\section{Conclusion}
Conditional anomaly detection is a promising methodology for detecting unusual events that may correspond to medical error or unusual outcomes. We have proposed a new anomaly detection approach that uses the discriminative projection techniques to identify anomalies. The method generalizes previously proposed probabilistic anomaly detection framework \cite{hauskrecht2007evidence-based}.  The advantage of the method is that it performs fully unsupervised and with the minimum input from the domain expert. 
The new method was tested on the new heparin--induced thrombocytopenia dataset with over 30000 patient state entries. The experiments demonstrated that our evidence-based anomaly detection methods can detect clinically important anomalies very well, with the detector based on the SVM projections reaching performance level of 15.6\%. These statistics are very encouraging since they compare favorably to performance rates achieved by the rule--based system currently employed in the hospital to detect the HIT condition.

\section*{Acknowledgments} 
 
This research was funded by grants R21--LM009102 and R01--LM008374 from the National Library of Medicine.


\bibliography{miki}
\bibliographystyle{mlapa}

\end{document}